\documentclass[10pt,twocolumn,letterpaper]{article}

\usepackage[pagenumbers]{wacv} 

\usepackage{algorithm}
\usepackage{algpseudocode}
\usepackage{graphicx}
\usepackage{amsmath}
\usepackage{amssymb}
\usepackage{booktabs}
\usepackage[utf8]{inputenc} 
\usepackage[T1]{fontenc}    
\usepackage{url}            
\usepackage{booktabs}       
\usepackage{amsfonts}       
\usepackage{nicefrac}       
\usepackage{microtype}      
\usepackage{xcolor}         
\usepackage{graphicx}
\usepackage{enumitem}
\usepackage{multicol, multirow}
\usepackage{makecell}
\usepackage{colortbl}
\definecolor{gray}{RGB}{230,230,230}
\usepackage{amsfonts}
\usepackage{bm}

\usepackage[capitalize]{cleveref}
\crefname{section}{Sec.}{Secs.}
\Crefname{section}{Section}{Sections}
\Crefname{table}{Table}{Tables}
\crefname{table}{Tab.}{Tabs.}

\def\wacvPaperID{133} 
\def\confName{WACV}
\def\confYear{2025}

\begin{document}

\title{GaussianSR: High Fidelity 2D Gaussian Splatting for
Arbitrary-Scale Image Super-Resolution}


\author{
    Jintong Hu$^{1*}$, Bin Xia$^{2*}$, Bin Chen$^{3*}$, Wenming Yang$^{1\dagger}$, Lei Zhang$^{4,5}$ \\
    $^1$Shenzhen International Graduate School, Tsinghua University \\
    $^2$Department of Computer Science and Engineering, The Chinese University of Hong Kong \\ 
    $^3$School of Electronic and Computer Engineering, Peking University \\
    $^4$The Hong Kong Polytechnic University \quad $^5$OPPO Research Institute
}

\maketitle

\begin{abstract}
  Implicit neural representations (INRs) have significantly advanced the field of arbitrary-scale super-resolution (ASSR) of images. Most existing INR-based ASSR networks first extract features from the given low-resolution image using an encoder, and then render the super-resolved result via a multi-layer perceptron decoder. Although these approaches have shown promising results, their performance is constrained by the limited representation ability of discrete latent codes in the encoded features. In this paper, we propose a novel ASSR method named GaussianSR that overcomes this limitation through 2D Gaussian Splatting (2DGS). Unlike traditional methods that treat pixels as discrete points, GaussianSR represents each pixel as a continuous Gaussian field. The encoded features are simultaneously refined and upsampled by rendering the mutually stacked Gaussian fields. As a result, long-range dependencies are established to enhance representation ability. In addition, a classifier is developed to dynamically assign Gaussian kernels to all pixels to further improve flexibility. All components of GaussianSR (i.e., encoder, classifier, Gaussian kernels, and decoder) are jointly learned end-to-end. Experiments demonstrate that GaussianSR achieves superior ASSR performance with fewer parameters than existing methods while enjoying interpretable and content-aware feature aggregations.
\end{abstract}

\footnote{This work is done during the internship at OPPO Research Institure.\\ $^{*}$Contributing equally to this work. $^{\dagger}$Corresponding Author: Wenming Yang, yang.wenming@sz.tsinghua.edu.cn \\}

\section{Introduction}
\label{sec:intro}

The vision world is continuous, presenting scenes and objects in their natural, uninterrupted forms. However, computer vision tasks predominantly employ pixel-based discrete representations for image processing, which inherently constrains the ability capture the continuous nature with high fidelity across different resolutions. The emergence of implicit neural representations (INR) has revolutionized this challenge by treating images as continuous functions. Pioneered by LIIF \cite{LIIF}, INR-based methods leverage Multi-Layer Perceptrons (MLPs) to model the RGB values of high-resolution (HR) images as continuous functions of low-resolution (LR) features and pixel coordinates. This continuous representation enables LIIF to perform super-resolution at arbitrary scales without the need for creating and training separate models for each scaling factor. Building upon LIIF, numerous variants and extensions have subsequently emerged \cite{LTE,xu2022ultrasr,ITSRN,cao2023ciaosr}, further advancing the capabilities of INR-based Arbitrary-Scale Super-Resolution (ASSR).

\begin{figure*}[t]
  \centering
  \includegraphics[width=1\linewidth]{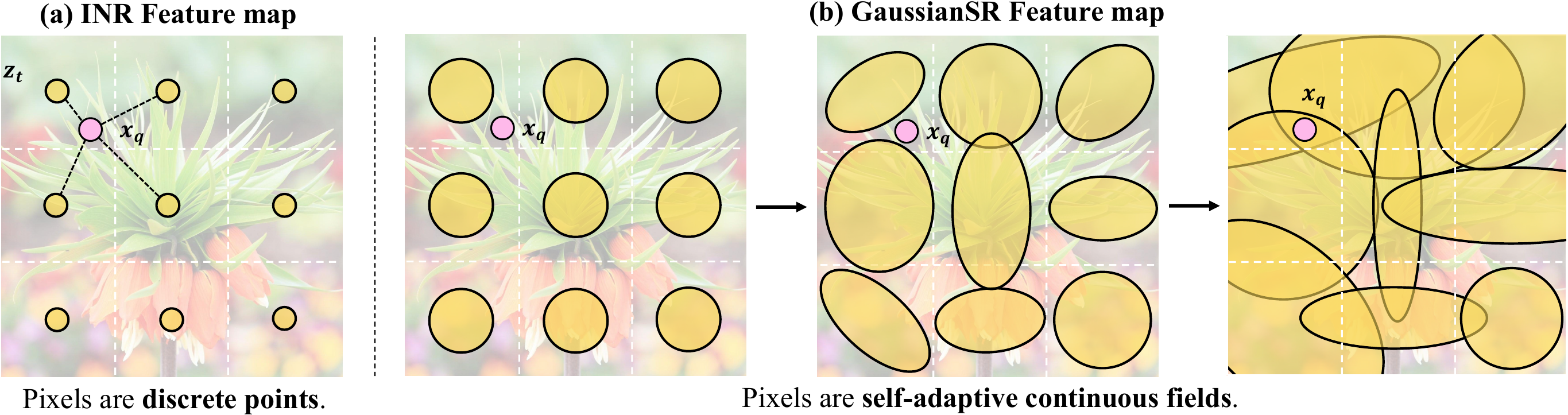}
   \caption{\textbf{Comparison of Feature Storage between INR-based ASSR and our GaussianSR.} INR methods treat pixels as discrete points. Instead, our GaussianSR method models each pixel as a continuous Gaussian field. By representing pixels as continuous fields instead of discrete points, GaussianSR can explicitly represent the field values at any position (e.g $x_{q}$). GaussianSR achieves arbitrary-scale upsampling in a more elegant and natural way.}
   \label{figure 1}
\end{figure*}

Although INR-based ASSR methods have demonstrated their effectiveness, several intrinsic limitations remain. Latent codes are stored individually, which necessitates the use of Multi-layer Perceptrons (MLPs) to interpolate these discrete features into a continuous output. This process may not fully preserve key physical properties of the scene, such as lighting and texture consistency, reducing the overall interpretability and physical fidelity of the generated images. Moreover, as depicted in Figure \ref{figure 1}(a), when query location $x_{q}$ moves in 2D domain, the selection of $z_{t}$ can abruptly switch from one to another if $x_{q}$ crosses the dashed lines, potentially leading to checkerboard artifacts in the output image. To counteract this, a strategy involving a weighted combination of nearby latent codes is employed, which, while effective, increases the computational load and may become a bottleneck for real-time applications.

\textbf{How about continuously storing latent codes?} After discussing the limitations of existing INR-based methods in handling 2D image super-resolution (SR) with continuous integrity, we explore the potential application of concepts derived from 3D reconstruction technologies to address these issues. Inspired by the recent advancements in 3D Gaussian Splatting (3DGS) for detailed scene rendering and object reconstruction \cite{3dgs,SplaTAM,GS-SLAM,Gaussian_SLAM}, we propose a method that adapts this technology to more efficiently handle continuous data representation. The key concept of 3DGS involves representing objects as 3D Gaussian fields, allowing for the construction of continuous structures with significantly reduced parameter requirements. By applying this method, overlapped Gaussian spheres can form various detailed shapes, offering a potential pathway to achieve higher fidelity in 2D image SR while maintaining computational efficiency.

In this paper, we introduce a paradigm shift for ASSR by proposing GaussianSR pipeline, which utilizes 2D Gaussian Splatting (2DGS) to overcome the inherent discontinuity of pixel-based INR methods. GaussianSR is built upon the insight that pixel values intrinsically exhibit intensity variations which can be more accurately captured through a continuous Gaussian representation. As illustrated in Figure \ref{figure 1}(b), each pixel under our methodology is modeled as a self-adaptive continuous field instead of a single value. This continuous representation naturally and explicitly yields the field value at any query position $x_{q}$, avoiding the information loss associated with discrete representations.

Specifically, our method enables the dynamic adaptation of kernels to match varied input qualities effectively. We train a classifier that assigns a Gaussian kernel to each input pixel. Rather than applying a global kernel or a fixed set of kernels, the classifier can tailor the kernel for each pixel based on its specific characteristics, resulting in a more adaptive and context-aware processing of the input data. Moreover, our technique overcomes the limitations of fixed receptive fields inherent in traditional INR-based ASSR methods by employing flexible and adaptable Gaussian kernels, which can be stacked together. This flexibility not only enables the model to capture features across multiple scales but also bolsters its capacity for representing complex, enhanced features, resulting in superior performance in SR tasks. Our efforts have successfully built a bridge between Gaussian representations and image restoration, marking a pioneering step in the field of ASSR. In summary, the contribution of this paper can be summarized into three parts:

\begin{itemize}
    \item We pioneer building a novel paradigm for ASSR through the 2D Gaussian Splatting representation, namely GaussianSR. Our approach builds a bridge between discrete and continuous feature representation by turning pixel point into Gaussian field.
    \item We train a classifier that adaptively assigns the appropriate 2D Gaussian kernel to each pixel, which facilitates the adaptive matching of diverse input characteristics with suitable Gaussian kernels, effectively accommodating a wide range of input variations.
    \item Extensive experiments demonstrate that seamlessly transitioning existing INR-based ASSR to our GaussianSR framework can achieve better performance with reduced parameter, underscoring the efficacy and great capability of our proposed approach.
\end{itemize}

\begin{figure*}[t]
  \centering
  \includegraphics[width=1\linewidth]{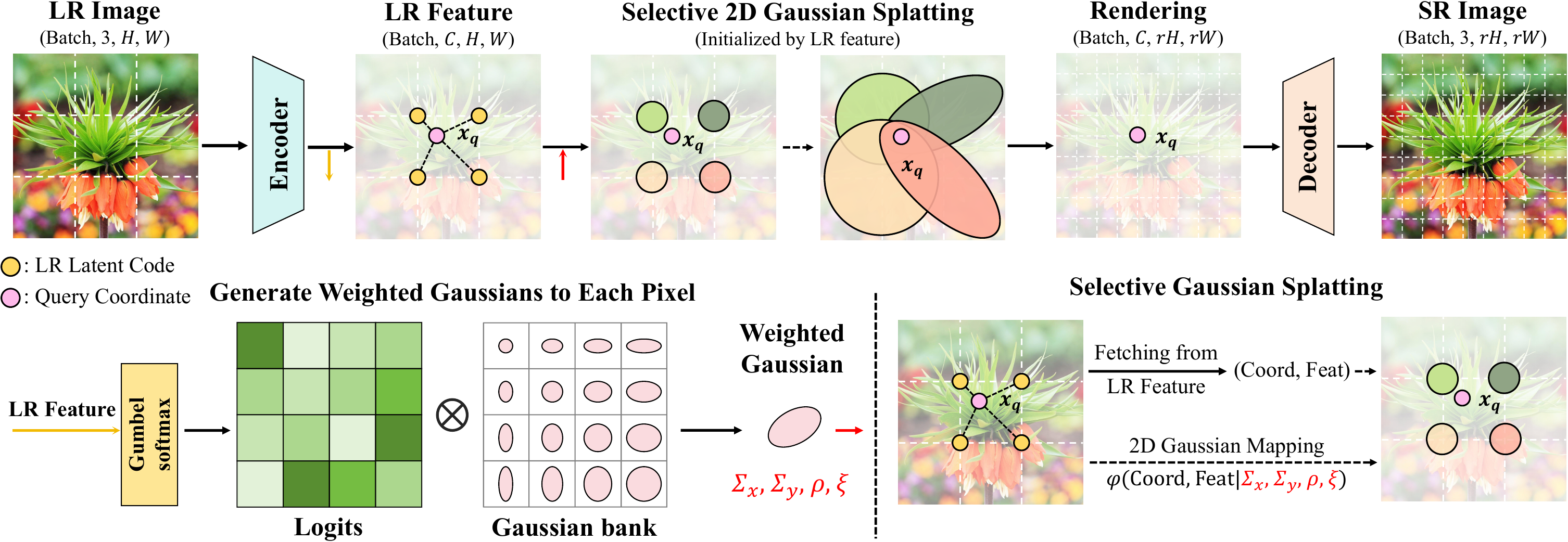}
   \caption{\textbf{The main pipeline of GaussianSR.} GaussianSR begins with an encoder that extracts feature representations from the input image, followed by Selective Gaussian Splatting which assigns a learnable Gaussian kernel to each pixel, converting dicrete feature points into Gaussian fields. Features at any arbitrary query point $x_{q}$ in the plane are computed using the overlapping Gaussian functions that modulate their influence based on the spatial location. Finally, these continuous-domain features are rendered into a high-resolution space and refined through the decoder to reconstruct the desired RGB output at the specified query coordinates.}
   \label{figure 2}
\end{figure*}

\section{Related Work}
\label{sec:related}
\subsection{Implicit Neural Representation}

Implicit Neural Representations (INRs) elegantly model signals across a continuous domain and have gained prominence in numerous fields, including 3D reconstruction \cite{DeepSDF,LIF,LIGR,Occupancy_Networks,PiFu}, scene rendering \cite{D-NERF,nerfies,nonrigid,Scene_Representation_Networks}, robotics \cite{chen2021fullbody,Li20213DNS,SE3}, and image and video compression \cite{chen2021nerv,strumpler2022inrcompress,zhang2022implicit}. Notably, NeRF \cite{NERF} innovatively represents complex 3D scenes with continuous neural implicit functions and has enabled novel view synthesis, although it faces challenges with dynamic scenes due to its static nature and lacks explicit geometric representations. Attempting to address dynamic content, Deformable Neural Radiance Fields \cite{nerfies} have introduced a warping mechanism to the static NeRF, and Neural Sparse Voxel Fields \cite{NSVF} have coupled implicit functions with sparse voxel grids to enhance efficiency and facilitate geometric interpretability. Despite the impressive outcomes of these methods, their discrete storage of each latent code or feature often misses the opportunity to encapsulate the intricacies or characteristics inherent to the signals in question. PixelNeRF \cite{pixelnerf} endeavors to overcome this by tying pixel-level features to the implicit representation, while HyperNeRF \cite{hypernerf} utilizes hypernetworks to augment the neural encoder’s capabilities, potentially allowing for a more nuanced capture of intrinsic properties.

\subsection{3D Gaussian Splatting}

3D Gaussian splatting (3DGS) \cite{3dgs} serves as a groundbreaking technique in computer graphics, offering an explicit scene representation through millions of learnable 3D Gaussians, a stark contrast to NeRF's \cite{NERF} implicit, coordinate-based approach. This state-of-the-art methodology harbors the promise of real-time rendering in addition to unparalleled degrees of controllability and editability. The practicality of 3DGS transcends the traditional confines, proving useful for simultaneous localization and mapping \cite{SplaTAM,GS-SLAM,Gaussian_SLAM}, dynamic scene modeling \cite{yang2023gs4d,wu20234dgaussians,luiten2023dynamic}, AI-generated content \cite{Zielonka2023Drivable3D,huang2023sc,chen2024textto3d}, and autonomous driving \cite{yan2024street,zhou2024drivinggaussian}. While 3DGS has significantly advanced 3D-related tasks, its application to image super-resolution had remained untapped before our work. Our pioneering work adapts the Gaussian Splatting to arbitrary-scale super-resolution, an area traditionally ruled by LIIF \cite{LIIF} and its variants \cite{xu2022ultrasr,ITSRN,LTE}, with our approach enhancing quality with reduced parameters. Our innovations set a path for integrating Gaussian representation into the broader field of image restoration.

\subsection{Arbitrary-Scale Super-Resolution}

\begin{figure*}
  \centering
  \includegraphics[width=\textwidth]{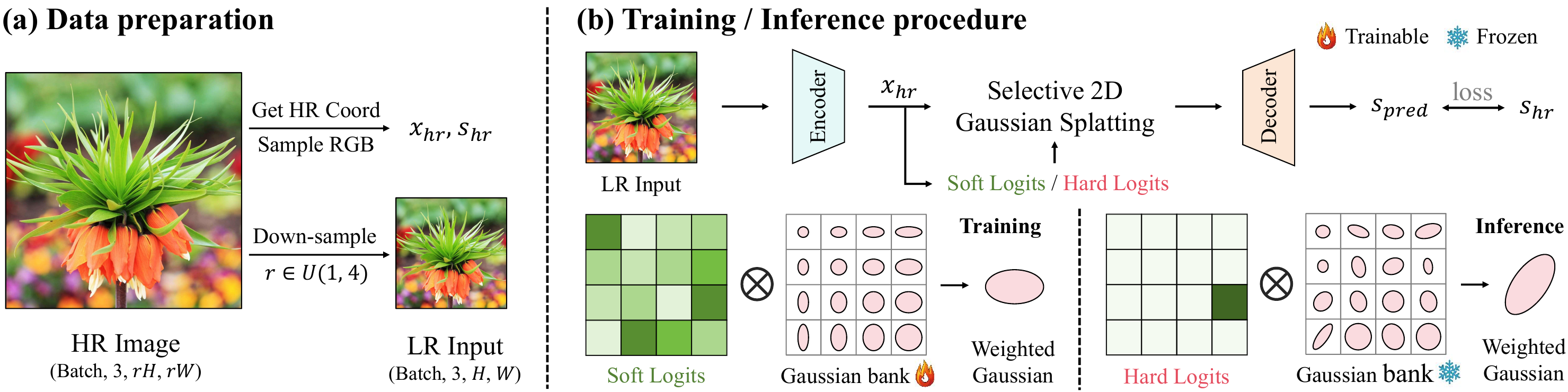}
  \caption{\textbf{Training and Inference Process of GaussianSR}. GaussianSR employs the Selective Gaussian Splatting (SGS) module, which adaptively assigns Gaussian kernels to pixels based on their distinctive features. During training, SGS leverages the Gumbel Softmax to generate soft labels, enabling gradient backpropagation and parameter optimization for both logits and the Gaussian bank (which stores standard deviations and opacities). In the inference phase, SGS switches to hard labels, selecting the most likely Gaussian kernel for each pixel based on the optimized parameters.}
  \label{figure 3}
\end{figure*}

Traditional single image super-resolution (SISR) techniques, such as interpolation, have been outperformed by deep learning-based methods, notably Convolutional Neural Networks (CNNs), due to their ability to learn hierarchical features and map LR to HR images. Various CNN architectures have been proposed, including sub-pixel convolutional layers, pyramid frameworks, memory blocks, residual blocks, and dense connections, etc \cite{SRCNN,EDSR,ESPCN,LapSRN,MemNet,RDN}. However, these methods lack flexibility as they are designed for fixed integer-scale upsampling, limiting their applicability.

Arbitrary-scale super-resolution (ASSR) techniques have gained prominence for their adaptability across various scaling factors. MetaSR \cite{MetaSR} pioneered CNN-based ASSR, while LIIF \cite{LIIF} introduced an innovative framework leveraging implicit neural representations to treat images as continuous functions. Subsequent studies, such as LTE \cite{LTE}, SADN \cite{SADN}, and A-LIIF \cite{A-LIIF}, have aimed to address limitations like spectral bias, multi-scale feature integration, and artifact mitigation. However, these methods treat all features discretely without accounting for inherent variations, lack adaptive receptive fields to naturally and explicitly handle inputs of diverse scales, and rely on generic kernel weights that cannot adapt to individual samples during inference.

\section{Model Architecture}
\label{sec:model}

Figure \ref{figure 2} depicts GaussianSR's architecture. Section \ref{3.1} reviews the theoretical foundation of 3D Gaussian Splatting (GS). Section \ref{3.2} proposes the 2DGS for image restoration based on 3DGS. Section \ref{3.3} introduces the arbitrary-scale super-resolution pipeline based on 2D Gaussian splatting. 

\subsection{Preliminary: 3D Gaussian Splatting}
\label{3.1}

3D Gaussian splatting (3DGS) \cite{3dgs} represents a paradigmatic shift from the prevalent neural radiance fields (NeRF) \cite{NERF} methodology for scene representation and rendering. Grounded in a fundamentally distinct approach, 3DGS circumvents the computational complexities and controllability limitations inherent to NeRF while retaining the capability to synthesize photorealistic novel views from sparse input data. The forward process of 3DGS can be summarized as: 
$\bullet$ \textbf{3D Gaussian Representation}. The scene is represented using a collection of 3D Gaussians, each characterized by learnable properties such as position, opacity, covariance matrix, and color. This explicit representation allows efficient rendering through parallelized workflows. 
$\bullet$ \textbf{Splatting and Tiling}. The 3D Gaussians are first projected onto the 2D image plane through a splatting process. The image is then divided into nonoverlapping patches or "tiles" to facilitate parallel computation. 
$\bullet$ \textbf{Sorted Gaussian Rendering}. The projected Gaussians are sorted by depth within each tile, and the final pixel color is computed by alpha compositing, leveraging the sorted order. 
The impact of a 3D Gaussian $i$ on an arbitrary 3D point p in 3D is defined as follows:
\begin{equation}
   f_i(p)=\sigma(\alpha_i)\exp(-\frac{1}{2}(p-\mu_i)\Sigma_i^{-1}(p-\mu_i))
\end{equation}
where $p$ represents an arbitrary point in a 3D Cartesian coordinate system. The 3D Gaussian i is parametrized by (1) mean $\mu_{i}$, (2) covariance $\Sigma_{i}$, (3) opacity $\sigma(\alpha_{i})$, (4) color parameters $c_{i}$, either 3 values for (R, G, B) or spherical harmonics coefficients. The image formation model of Gaussian splatting can be formulated as:
\begin{equation}
    C_{3DGS}(p)=\sum_{i \in N} c_if_i^{2D}(p)\prod_{j=1}^{i-1}(1-f_j^{2D}(p))
\end{equation}
where $f_i^{2D}$ is a projection of $f_i(p)$ into 2D, \ie onto an image plane of the camera that is being rendered. 


\subsection{2D Gaussian Splatting}
\label{3.2}

Although 3DGS \cite{3dgs} has shown remarkable performance in various 3D tasks, its inherent 3D formulation may not be optimally suited for certain 2D tasks such as image restoration. Motivated by the need for continuous representation, we propose a 2D adaptation of Gaussian splatting, termed 2D Gaussian Splatting (2DGS). Our 2DGS simplifies the approach compared to 3DGS by eliminating operations and parameters specific to 3D scene rendering, such as project transformations and spherical harmonics, streamlining the overall process and reducing complexity.


The process of 2DGS consists of two main steps: $\bullet$ \textbf{Grid Transformation}. Initially, we create an affine-transformed grid that takes into account the standard deviations of all Gaussian kernels. This grid is tailored to match the high resolution (HR) dimensions of the target output, preparing the scene for subsequent feature mapping. $\bullet$ \textbf{Feature Value Rendering}. Following the grid setup, the feature values undergo alpha blending, and then these blended values are projected onto the transformed grid for rendering. This step ensures that the upsampled image preserves natural continuity and visual integrity.

\begin{figure*}
  \centering
  \includegraphics[width=\textwidth]{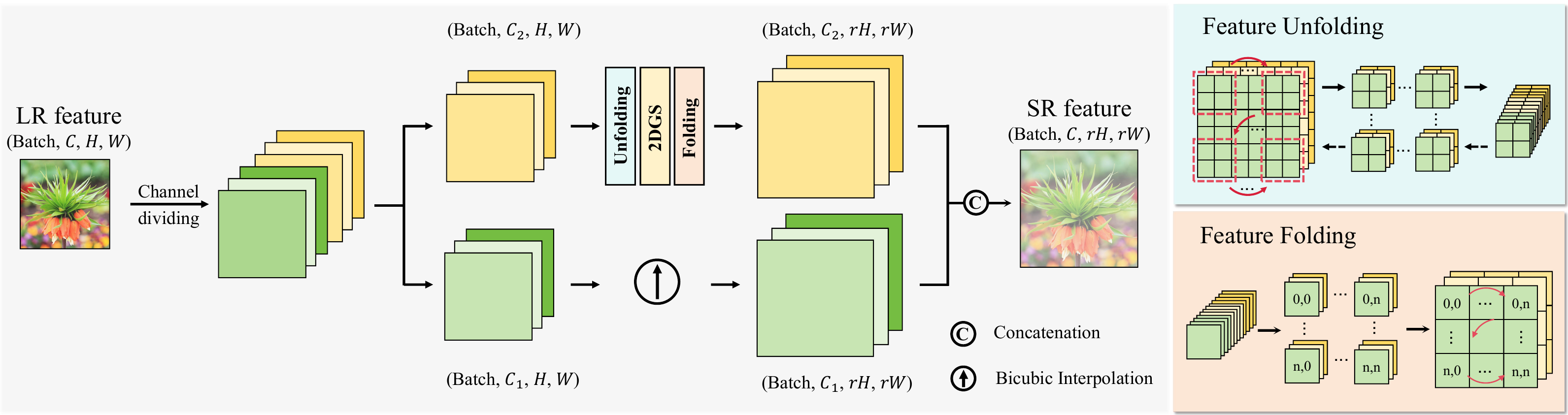}
  \caption{\textbf{The architecture of Dual-Stream Feature Decoupling}. The encoded features are decoupled along the channel dimension into two tensors, with one tensor undergoing feature unfolding, Gaussian splatting, and feature folding to preserve representational details, while the other tensor is bicubically upsampled for efficiency. The upsampled outputs from these two parallel streams are then fused back to the original channel.}
  \label{figure 4}
\end{figure*}

In our framework, the image representation unit is represented as a 2D Gaussian defined by its position $\mathbf{\mu}_{i} \in \mathbb{R}^{2}$ and the covirance matrix $\mathbf{\Sigma}_{i}=([\Sigma_{xi}, \rho_{i}], [\rho_{i}, \Sigma_{yi}]) \in \mathbb{R}^{2\times2}$. To prevent $\Sigma_x$, $\Sigma_y$ from being negative, we apply the sigmoid function for activation. The Gaussian field is represented as:
\begin{equation}
    \label{eq3}
    f_{i}(p|\mathbf{\mu}_{i}, \mathbf{\Sigma}_{i}) = \frac{1}{2\pi\left|\mathbf{\Sigma_{i}}\right|} e^{-\frac{1}{2}(p-\mathbf{\mu}_{i})^{\mathsf{T}}\mathbf{\Sigma_{i}}^{-1}(p-\mathbf{\mu}_{i})}
\end{equation}
where $p$ is an arbitrary coordinate in $\mathbb{R}^{2}$ space. In situations where multiple Gaussian fields overlap, we use alpha blending to combine their feature values $\mathbf{v}_{i} \in \mathbb{R}^{k}$, modulated by corresponding opacities $\mathbf{\xi}_{i} \in \mathbb{R}$: 
\begin{equation}
    \label{eq4}
    c_{i} = \sigma(\mathbf{\xi}) \cdot \mathbf{v}_{i}
\end{equation}

Based on the relative distance between the query coordinate $p$ and the center of each Gaussian field, we can determine the transmittance value $f_{i}(p|\mathbf{\mu}_{i}, \mathbf{\Sigma}_{i})$, which is then used to calculate a weighted sum with $c_i$, allowing us to render the value at $p$.
    


\subsection{Arbitrary-Scale Super-Resolution Pipeline}
\label{3.3}

The shift from a discrete feature storage methodology to a continuous feature field, characterized by a Gaussian distribution, represents a significant advancement in the field of image restoration. This transformation enables any point within the feature field to be explicitly determined according to the Gaussian distribution, thereby facilitating a more natural implementation of ASSR. Based on 2DGS, we have designed a novel image SR framework called GaussianSR, which introduces Gaussian representation for the first time in image restoration tasks. In this section, we discuss the key components of the GaussianSR framework.

\textbf{Overall Architecture}. \quad The architecture of GaussianSR is illustrated in Figure \ref{figure 2}. The process begins with an encoder that extracts features from the input images. These features are then mapped onto a learnable Gaussian field via Selective Gaussian Splatting, assigning a Gaussian distribution to each pixel. At any given query coordinate $x_{q}$, multiple overlapping Gaussian fields determine the feature value through their collective Gaussian functions explicitly. The features are subsequently rendered into the high-resolution space and processed through several fully connected layers to restore the channel dimension and produce the RGB value at $x_{q}$. Since $x_{q}$ can be any point within the $\mathbb{R}^{2}$ space, the framework achieves arbitrary-scale super-resolution.

\textbf{LR Initialization}. \quad LR feature initialization involves fetching features from the input LR image and representing them as initialization points within a continuous Gaussian feature field. Specifically, the value of the LR feature is used as the initial amplitude of the Gaussian, while the coordinates of the LR feature determine the center of the Gaussian field. Mathematically, let {($m_i$, $n_i$, $\mathbf{v}_i$)} be the set of LR feature coordinates and their corresponding values, where $i=1,2,\cdots,N$, and $N$ is the total number of features. Each LR feature is then represented as a Gaussian distribution $f_{i}(p|\mathbf{\mu}_{i}, \mathbf{\Sigma}_{i})$ within the continuous feature field, with its mean $\mathbf{\mu}_{i}$ = ($m_i$, $n_i$) corresponding to the normalized LR feature coordinates and its amplitude initialized to $\mathbf{v}_{i}$, the actual value of the feature of the LR image. The feature value at any coordinate $p$ within the continuous Gaussian feature field is determined by summing the function values from all individual Gaussian distributions centered at the LR feature coordinates:

\begin{figure*}
  \centering
  \includegraphics[width=\textwidth]{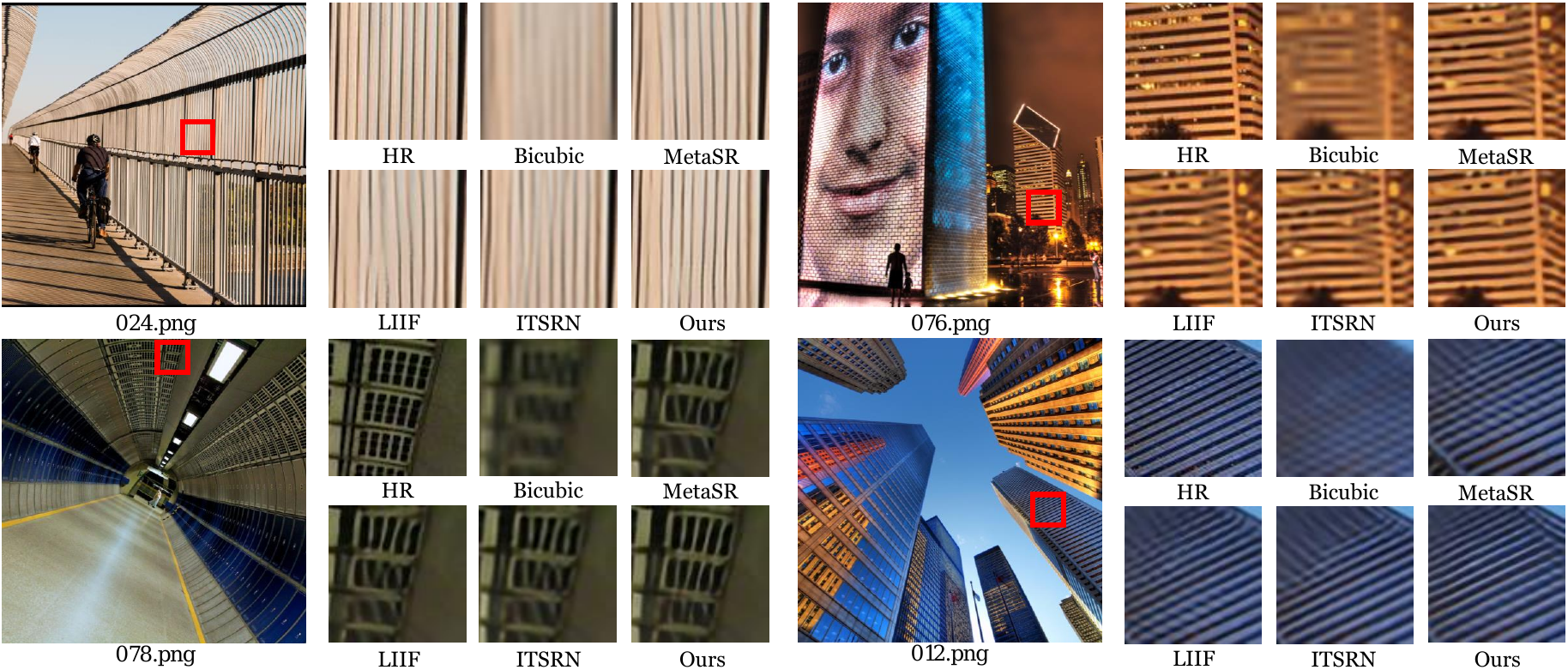}
  \caption{\textbf{Qualitative comparison for ×4 SR to other ASSR methods on Urban100 \cite{urban100} dataset}. EDSR-baseline \cite{EDSR} is used as an encoder for all methods. For all the shown examples, our method significantly outperforms other methods, particularly in the image rich in repeated textures and structures.}
  \label{figure 5}
\end{figure*}

\begin{equation}
     C_{2DGS}(p|\mathbf{v},\mathbf{\mu},\mathbf{\Sigma},\mathbf{\xi})=\sum_{i}f_{i}(p|\mathbf{\mu}_{i},\mathbf{\Sigma}_{i}) \cdot c_{i}  
\end{equation}
where $f_{i}(\cdot)$, $c_{i}(\cdot)$ are the same as defined in Equations \ref{eq3} amd \ref{eq4}. LR feature initialization ensures that subsequent upsampling and reconstruction processes can effectively leverage the spatial and intensity information from the LR image, ultimately leading to high-quality HR image reconstruction.

\textbf{Selective Gaussian Splatting}. \quad Convolutional operations, while effective, are inherently limited by their fixed kernel sizes, incapable of adapting to varying input characteristics. Moreover, high-resolution LR images can induce redundancy among Gaussian kernels due to similar standard deviations and opacities. To address these limitations, we propose the Selective Gaussian Splatting (SGS) technique, which adaptively assigns Gaussian kernels to pixels based on their unique features, minimizing kernel redundancy and parameter overhead. SGS leverages logits from an encoder corresponding to 100 distinct classes per pixel, each associated with a Gaussian kernel differentiated by standard deviation and opacity. Initially varied in shape and transparency (refer to Figure \ref{figure 3}), these kernels are optimized during training, aligning with encoder logits to generate customized Gaussian distributions for each pixel location. Gumbel Softmax facilitates gradient backpropagation with soft labels during training and hard labels during inference, enhancing optimization while maintaining interpretability. By selectively assigning distinct kernels based on pixel-wise characteristics, SGS markedly improves flexibility in modeling pixel distributions, thereby enhancing performance across computer vision tasks while reducing parameter overhead.

\textbf{Dual-Stream Feature Decoupling}. \quad Addressing the issues of high memory consumption inherent in Gaussian splatting methodologies, we introduce the Dual-Stream Feature Decoupling architecture, specifically aimed at enhancing super-resolution processes while maintaining parameter efficiency. This approach strategically decouples encoded features along the channel dimension into two tensors with reduced channels but preserved spatial resolution (refer to Figure \ref{figure 4}). For one tensor, an additional feature unfolding step is performed before Gaussian splatting. This unfolding operation rearranges the input tensor into a lower-resolution representation, effectively reducing the memory consumption for the subsequent Gaussian splatting operation. After Gaussian splatting for detail preservation, a corresponding feature folding operation restores the original spatial dimensions. The other tensor stream is bicubically upsampled for efficiency. The augmented outputs from these two parallel streams are then fused and refined through the decoder, producing the high-resolution RGB output. 

\section{Experiments}
\label{sec:experiments}

\begin{table*}[!h]
    \centering
    \footnotesize
    \setlength{\tabcolsep}{4pt}
    \renewcommand{\arraystretch}{1.1}
    \caption{Quantitative comparison for integer-scale super-resolution on other benchmarks (PSNR (dB)). The best results for each case are in \textbf{bold}. $^{\dagger}$ As the DIINN method has not been open-sourced, we directly use the results reported in their article.}
    \begin{tabular}{l||c c c|c c c|c c c|c c c|c c c}
    \Xhline{1pt}
        \rowcolor{gray} ~ & \multicolumn{3}{c|}{General100} & \multicolumn{3}{c|}{BSD100} & \multicolumn{3}{c|}{Urban100} & \multicolumn{3}{c|}{Manga109} & \multicolumn{3}{c}{DIV2K100} \\ \Xcline{2-16}{0.01pt} 
        \rowcolor{gray}
        \multirow{-2}{*}{Methods} & ×2 & ×3 & ×4 & ×2 & ×3 & ×4 & ×2 & ×3 & ×4 & ×2 & ×3 & ×4 & ×2 & ×3 & ×4 \\ \hline\hline
        Bicubic & 32.14 & 28.56 & 26.58 & 28.25 & 25.96 & 24.69 & 25.68 & 23.07 & 21.77 & 29.98 & 25.68 & 23.52 & 31.45 & 28.42 & 26.81 \\
        EDSR-baseline & 38.23 & 33.93 & 31.48 & 32.16 & 29.09 & 27.57 & 31.98 & 28.15 & 26.04 & 38.54 & 33.45 & 30.35 & 34.55 & 30.90 & 28.94 \\ 
        EDSR-baseline-MetaSR & 38.22 & 33.93 & 31.40 & 32.16 & 29.09 & 27.55 & 32.08 & 28.12 & 25.95 & 38.53 & 33.51 & 30.37 & 34.64 & 30.93 & 28.92 \\
        EDSR-baseline-LIIF & 38.25 & 33.97 & 31.53 & 32.17 & 29.10 & 27.60 & 32.15 & 28.22 & 26.15 & 38.63 & 33.47 & 30.54 & 34.67 & 30.96 & 29.00 \\
        EDSR-baseline-ITSRN & 38.25 & 33.95 & 31.48 & 32.18 & 29.10 & 27.58 & 32.13 & 28.14 & 26.06 & 38.58 & 33.47 & 30.47 & 34.67 & 30.93 & 28.97 \\
        EDSR-baseline-ALIIF & 38.21 & 33.95 & 31.48 & 32.18 & 29.11 & 27.60 & 32.09 & 28.19 & 26.14 & 38.53 & 33.42 & 30.47 & 34.65 & 30.95 & 28.99 \\
        EDSR-baseline-DIINN$^{\dagger}$ & - & - & - & 30.69 & 27.73 & 26.22 & 30.29 & 26.46 & 24.49 & - & - & - & 34.63 & 30.93 & 28.98 \\
        EDSR-baseline-GaussianSR & \textbf{38.31} & \textbf{34.02} & \textbf{31.55} & \textbf{32.20} & \textbf{29.13} & \textbf{27.61} & \textbf{32.25} & \textbf{28.28} & \textbf{26.19} & \textbf{38.64} & \textbf{33.57} & \textbf{30.54} & \textbf{34.71} & \textbf{31.00} & \textbf{29.03}\\ 
    \Xhline{1pt}
    \end{tabular}
    
    \label{table1}
\end{table*}

\begin{table*}[!h]
    \centering
    \footnotesize
    \setlength{\tabcolsep}{4.1pt}
    \renewcommand{\arraystretch}{1.1}
    \caption{Quantitative comparison for arbitrary noninteger-scale super-resolution on other benchmarks (PSNR (dB)). The best results for each case are in \textbf{bold}. $^{\dagger}$ EDSR-baseline \cite{EDSR} cannot handle noninteger-scale super-resolution.}
    \begin{tabular}{l||c c c|c c c|c c c|c c c|c c c}
    \Xhline{1pt}
        \rowcolor{gray} ~ & \multicolumn{3}{c|}{General100} & \multicolumn{3}{c|}{BSD100} & \multicolumn{3}{c|}{Urban100} & \multicolumn{3}{c|}{Manga109} & \multicolumn{3}{c}{DIV2K100} \\ \Xcline{2-16}{0.01pt} 
        \rowcolor{gray}
        \multirow{-2}{*}{Methods} & ×1.5 & ×2.4 & ×3.6 & ×1.5 & ×2.4 & ×3.6 & ×1.5 & ×2.4 & ×3.6 & ×1.5 & ×2.4 & ×3.6 & ×1.5 & ×2.4 & ×3.6 \\ \hline\hline
        Bicubic & 34.89 & 30.12 & 27.17 & 30.78 & 27.09 & 25.11 & 27.92 & 24.25 & 22.21 & 33.12 & 27.50 & 24.15 & 34.00 & 29.75 & 27.31 \\
        EDSR-baseline$^{\dagger}$ & - & - & - & - & - & - & - & - & - & - & - & - & - & - & - \\
        EDSR-baseline-MetaSR & 42.16 & 36.14 & 32.30 & 35.69 & 30.60 & 28.08 & 36.05 & 30.07 & 26.74 & 42.67 & 36.18 & 31.56 & 38.57 & 32.78 & 29.62 \\
        EDSR-baseline-LIIF & 42.20 & 36.18 & 32.37 & 35.69 & 30.62 & 28.11 & 36.13 & 30.16 & 26.89 & 42.67 & 36.19 & 31.59 & 38.57 & 32.82 & 29.67 \\
        EDSR-baseline-ITSRN & 42.24 & 36.18 & 32.34 & 35.70 & 30.62 & 28.10 & 36.14 & 30.12 & 26.81 & 42.68 & 36.18 & 31.55 & 38.61 & 32.80 & 29.64 \\
        EDSR-baseline-ALIIF & 42.14 & 36.16 & 32.33 & 35.67 & 30.60 & 28.10 & 36.02 & 30.07 & 26.84 & 42.56 & 36.11 & 31.53 & 38.54 & 32.79 & 29.65 \\
        EDSR-baseline-GaussianSR & \textbf{42.24} & \textbf{36.23} & \textbf{32.40} & \textbf{35.73} & \textbf{30.64} & \textbf{28.13} & \textbf{36.27} & \textbf{30.23} & \textbf{26.94} & \textbf{42.72} & \textbf{36.25} & \textbf{31.60} & \textbf{38.64} & \textbf{32.85} & \textbf{29.70}\\ 
    \Xhline{1pt}
    \end{tabular}
    
    \label{table2}
\end{table*}

\subsection{Datasets}

For our training, we use the DIV2K dataset \cite{DIV2K}. Sourced from the NTIRE challenge \cite{NTIRE}, this dataset comprises 1000 diverse 2K-resolution images featuring a wide range of content, including individuals, urban scenes, flora, fauna and natural landscapes. Within this collection, we allocate 800 images for the training set, 100 images for validation. To evaluate the generalization performance of the model, we report the results on the DIV2K validation set with 100 images. Another four benchmark datasets are also utilized, namely General100 \cite{General100}, BSD100 \cite{B100}, Urban100 \cite{urban100}, and Manga109, \cite{Manga109} which provided a comprehensive landscape for assessing cross-dataset robustness. Consistent with previous work \cite{LIIF,MetaSR,ITSRN}, we utilized bicubic downsampling to synthesize the LR images.

\subsection{Implementation details}

To simulate a continuous magnification process, the downsampling factor is randomly sampled from a uniform distribution, U(1, 4), enabling the model to adapt to different degrees of image degradation. To accommodate the GaussianSR framework, within each batch, the downsampling factor remains constant across different GPUs (if multi-GPU training is employed). The loss function used is the $L_{1}$ distance between the reconstructed image and the ground truth image. Following the settings in LIIF and its variants \cite{LIIF,A-LIIF,ITSRN}, we randomly crop the LR images into 48$\times$48 patches, collect 2304 random pixels on the HR images. The initial learning rate for all modules is set to 1e-4, and is halved every 200 epochs. The model is trained in parallel on 4 Tesla V100 GPUs with a mini-batch size of 16. The training process takes about 2000 epochs to converge. 


\subsection{Quantitative and Qualitative Results}

To validate the efficacy of our GaussianSR, we perform a comparative analysis against several advanced methods. These methods include MetaSR \cite{MetaSR}, a pioneering work that first realizes super-resolution for non-integer scales using a CNN-based method. Additionally, we have chosen LIIF \cite{LIIF}, which introduced the concept of implicit neural representations to the image super-resolution domain. Furthermore, we have included several variants and extensions of the LIIF method in our comparison, such as ITSRN \cite{ITSRN}, A-LIIF \cite{A-LIIF}, DIINN \cite{DIINN}. We re-train these models under the same framework to ensure fairness. Due to the significantly larger number of parameters and the different framework employed by CiaoSR \cite{cao2023ciaosr}, it was not included in the comparative analysis. Tables \ref{table1} and \ref{table2} present the results for integer-scale SR and noninteger-scale SR, respectively. Our method achieves competitive results across all 5 datasets for all scaling factors, particularly excelling in noninteger scaling factors due to the flexible representation of Gaussians. Although our model performs comparably to LIIF and A-LIIF on lower-resolution datasets such as General100 and BSD100, it shows significant improvement on higher-resolution datasets like Urban100 and Manga109, with particularly notable results on Urban100.

Figure \ref{figure 5} offers a qualitative comparison with other arbitrary-scale SR methods. Our model excels at synthesizing SR images with sharper textures. For instance, in the second column of the first row and the second column of the second row, our method successfully recovers the texture of buildings despite significant texture degradation in the low-resolution images. In contrast, other methods tend to produce erroneous redundant artifacts that detract from the visual quality of the image due to limitations in local integration and insufficient feature extraction. More visual comparisons can be found in the supplementary materials.

In Table \ref{table:3}, we present a comparative analysis of our method against LIIF and ITSRN \cite{ITSRN} at early stage (epoch 100). Our method consistently outperforms other methods on Urban100 and BSD100, underscoring its superior performance even before full training is completed. This shows that our approach is more effective in leveraging a limited amount of training samples to achieve better results. Notably, we ensure that LIIF and ITSRN are trained under the same framework for a fair comparison. The results clearly indicate that our method is capable of achieving enhanced performance with fewer training samples. In addition, we report the inference time of GaussianSR and LIIF with the same input and output sizes in Table \ref{table:4}, providing a comprehensive evaluation of their efficiency.

\subsection{Ablation Study}

In this section, we conduct an ablation study to investigate the effects of different settings in our architecture. Table \ref{table:5} and Table \ref{table:6} present the results of various hyperparameter settings for channel decoupling and Gaussian bank.

\textbf{Channel decoupling.} \quad The default configuration involves 8 channels undergoing 2D Gaussian Splatting (2DGS) upsampling, while the remaining channels are upsampled using bicubic interpolation. This setup is designed to achieve memory-efficient training and inference. We compare this configuration with settings where all channels were upsampled using bicubic interpolation and where 16 channels were upsampled using 2DGS. The version with 8 channels using 2DGS achieves the highest PSNR on the DIV2K validation set, and all versions with 2DGS significantly outperformed the version using only bicubic interpolation. These results demonstrate the effectiveness of our proposed framework for super-resolution tasks.

\begin{table}
    \centering
    \footnotesize
    \setlength{\tabcolsep}{7.2pt}
    \renewcommand{\arraystretch}{1.1}
    \caption{Latency comparison in milliseconds. The input is a single RGB image of size 64$\times$64. We report the average runtime over 100 runs on a Tesla V100 GPU. Throughout the following tables, the best results for each case are in \textbf{bold}.}
    \begin{tabular}{c|c c c}
    \Xhline{1pt}
        \rowcolor{gray} Target Resolution & (128$\times$128) & (256$\times$256) & (384$\times$384) \\ \Xcline{1-4}{0.01pt} \rowcolor{gray}
        Methods & \multicolumn{3}{c}{Runtime (ms) $\downarrow$} \\ \hline\hline 
        LIIF & 14.52 & 43.18 & \textbf{86.43} \\
        ITSRN & 21.55 & 58.01 & 120.69 \\
        GaussianSR & \textbf{12.56} & \textbf{41.70} & 106.72 \\
    \Xhline{1pt}
    \end{tabular}
    
    \label{table:3}
\end{table}

\begin{table}
    \centering
    \footnotesize
    \setlength{\tabcolsep}{6pt}
    \renewcommand{\arraystretch}{1.1}
    \caption{Comparative analysis of early stage performance between GaussianSR and LIIF on Urban100 and BSD100. LIIF is trained under the same conditions to ensure a fair comparison.}
    \begin{tabular}{c|c c c | c c c}
    \Xhline{1pt}
        \rowcolor{gray} Epoch & \multicolumn{3}{c|}{Urban100} & \multicolumn{3}{c}{BSD100} \\ \Xcline{1-7}{0.01pt} \rowcolor{gray}
        Methods & x2 & x3 & x4 & x2 & x3 & x4 \\ \hline\hline 
        LIIF & 31.35 & 27.62 & 25.65 & 31.98 & 28.95 & 27.44 \\
        ITSRN & 31.45 & 27.54 & 25.56 & 32.03 & 28.94 & 27.43 \\ 
        GaussianSR & \textbf{31.63} & \textbf{27.79} & \textbf{25.78} & \textbf{32.06} & \textbf{29.00} & \textbf{27.50} \\
    \Xhline{1pt}
    \end{tabular}
    
    \label{table:4}
\end{table}

\begin{table}
    \centering
    \footnotesize
    \setlength{\tabcolsep}{8pt}
    \renewcommand{\arraystretch}{1.1}
    \caption{Ablation study on the channel proportion of Gaussian Splatting in Dual-Stream Feature Decoupling. (PSNR (dB)).}
    \begin{tabular}{l|c c c}
    \Xhline{1pt}
        \rowcolor{gray} ~ & \multicolumn{3}{c}{DIV2K100} \\ \Xcline{2-4}{0.01pt} \rowcolor{gray}
        \multirow{-2}{*}{Setting} & ×2 & ×3 & ×4 \\ \hline\hline 
        Bicubic-64 & 34.4128 & 30.8311 & 28.8788 \\
        Bicubic-56, Gaussian-8 & \textbf{34.7134} & \textbf{31.0010} & \textbf{29.0281} \\
        Bicubic-48, Gaussian-16 & 34.6990 & 30.9853 & 29.0157 \\
    \Xhline{1pt}
    \end{tabular}
    
    \label{table:5}
\end{table}

\begin{table}
    \centering
    \footnotesize
    \setlength{\tabcolsep}{7.2pt}
    \renewcommand{\arraystretch}{1.1}
    \caption{Ablation study on the number of Gaussian ellipses in Gaussian Bank. (PSNR (dB)).}
    \begin{tabular}{c|c c c}
    \Xhline{1pt}
        \rowcolor{gray} ~ & \multicolumn{3}{c}{Scale Factor} \\ \Xcline{2-4}{0.01pt} \rowcolor{gray}
        \multirow{-2}{*}{Classes of Gaussians} & ×2 & ×3 & ×4 \\ \hline\hline 
        100 & \textbf{34.7134} & \textbf{31.0010} & \textbf{29.0281} \\
        400 & 34.7041 & 30.9914 & 29.0174 \\
        900 & 34.7072 & 30.9906 & 20.0229 \\
    \Xhline{1pt}
    \end{tabular}

    \label{table:6}
\end{table}

\begin{figure}
  \centering
  \includegraphics[width=0.4\textwidth]{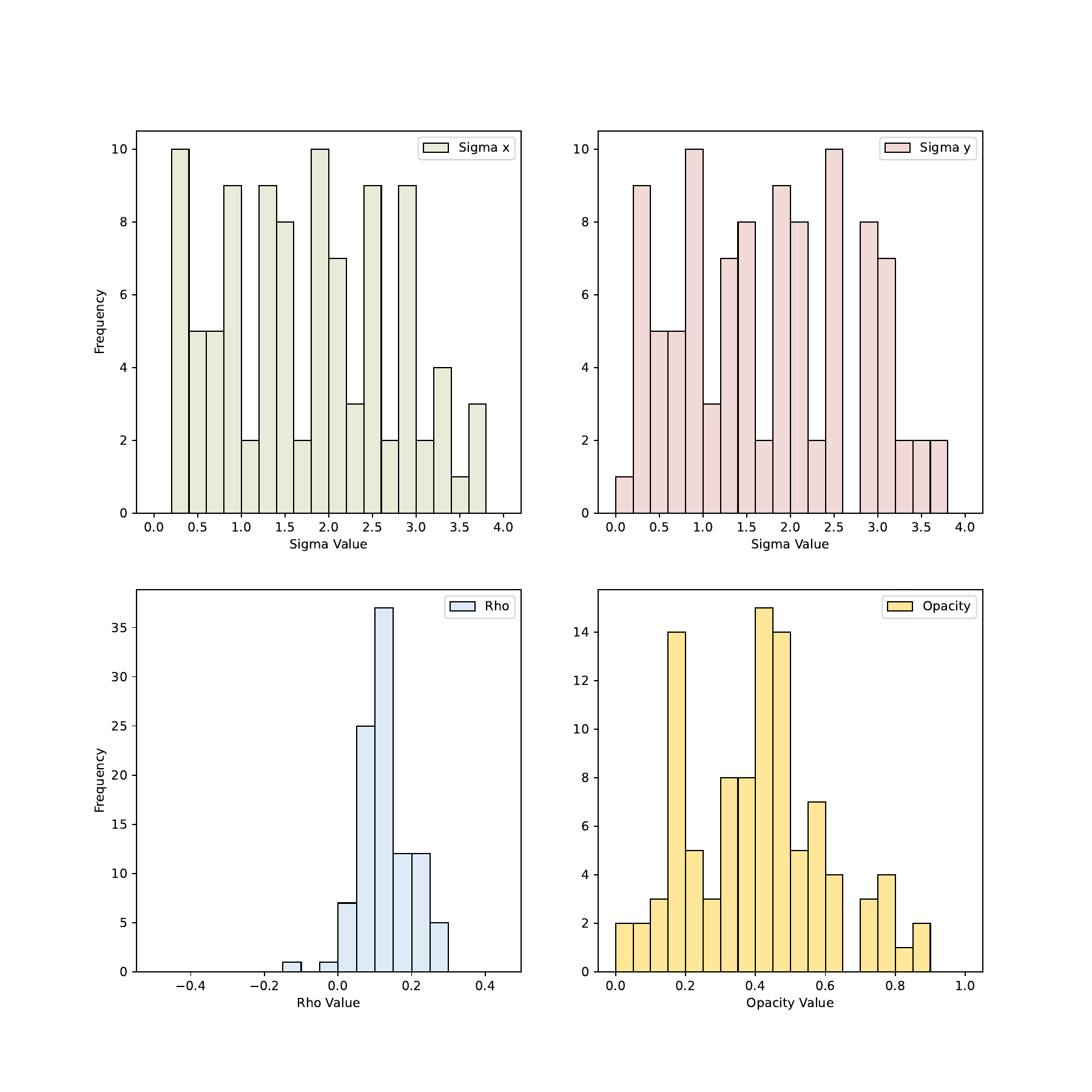}
  \caption{\textbf{Distributions of sigma $x$, simga $y$, $\rho$, and opacity in GaussianSR}. The final indicate that the majority of sigma values are significantly greater than zero, demonstrating that each pixel is represented by a Gaussian ellipse with a finite area. This finding validates the approach of replacing points with fields.}
  \label{figure 6}
\end{figure}

\textbf{Gaussian bank.} \quad GaussianSR consists of 100 Gaussian fields initialized in a stepwise manner. We conducted ablation experiments by varying the number of Gaussian fields to 100, 400, and 900. Interestingly, the configuration with 100 Gaussian produces better results. In the supplementary materials, we report the frequency of selection for each Gaussian, showing that fewer than 20 Gaussian ellipses were frequently selected. Thus, increasing the number of Gaussian fields is detrimental to training. Figure \ref{figure 6} visualizes the parameters within the Gaussian bank, showing that the majority of sigma values are significantly greater than zero at the end of training. This indicates that each pixel is represented by an ellipse with a finite area, validating our approach of replacing points with fields.

\section{Discussion}
\textbf{Discussion of Selective Gaussian Splatting.} \quad Convolutional operations are limited by their fixed kernel sizes and can induce redundancy in high-resolution images. Selective Gaussian Splatting technique adaptively assigns Gaussian kernels to pixels based on their unique features. This method minimizes kernel redundancy and parameter overhead. SGS uses logits from an encoder to assign one of 100 distinct Gaussian kernels to each pixel, optimized during training using Gumbel Softmax to facilitate gradient backpropagation. This adaptive approach improves flexibility and performance in modeling pixel distributions for computer vision tasks.

\begin{figure}[t]
  \centering
  \includegraphics[width=1\linewidth]{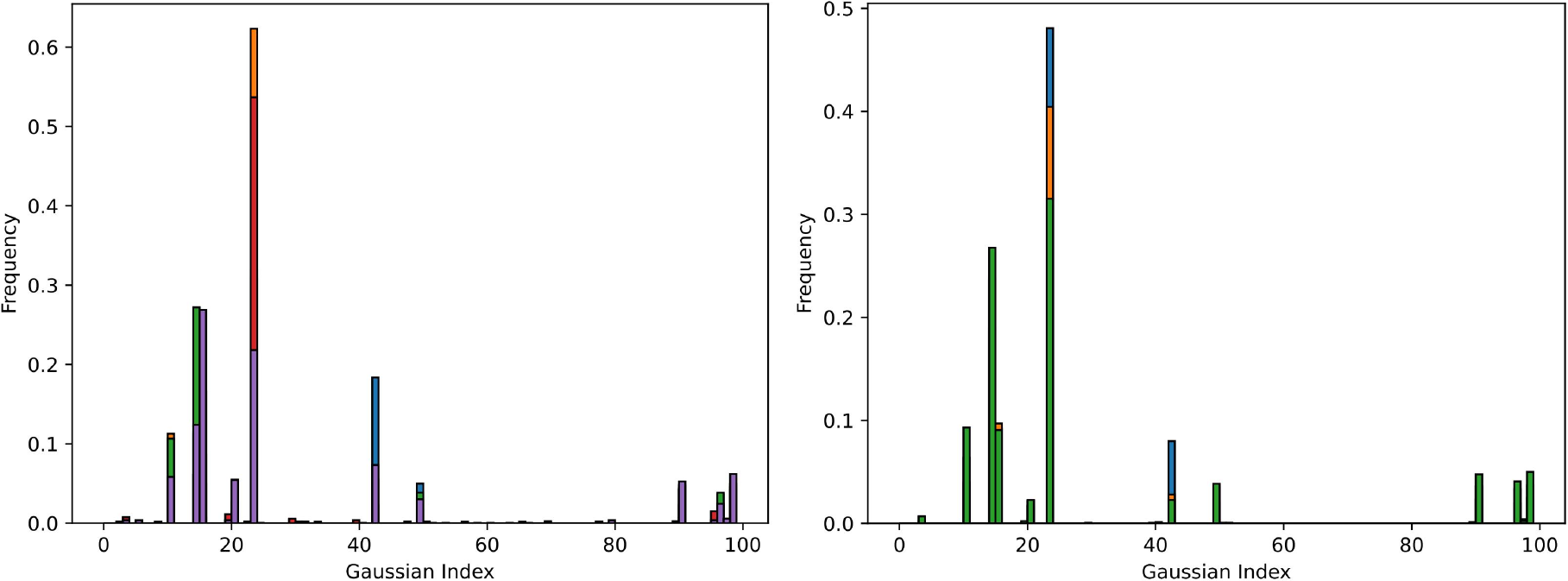}
   \caption{(a) Frequency of Gaussian kernel selection for five images from the Set5 \cite{set5} dataset under a 2$\times$ super-resolution setting. (b) Frequency of Gaussian kernel selection for the same image from the Set5 dataset under different scaling factors.}
   \label{figure 7}
\end{figure}

We visualize the logits results to demonstrate the frequency of selection for each Gaussian kernel. We conduct experiments on the Set5 dataset \cite{set5}, considering both different images with the same scaling factor and the same image with different scaling factors. Figure \ref{figure 7} (a) illustrates the frequency distribution for five images in Set5 under 2x super-resolution. The results show significant differences in frequency among all five images. Considering the high variance within Set5, we further experiment with the same image under different scaling factors. The results in Figure \ref{figure 7} (b) indicate significant differences in frequency as well, validating the correctness and effectiveness of our adaptive Gaussian kernel assignment approach.

In order to investigate the necessity of utilizing LR features, we conduct an overfitting analysis on a single image using the 2D Gaussian Splatting approach. The results reveal that the number of Gaussian fields significantly impacts the final reconstruction quality. Consequently, to fully leverage the information from the LR features, we represent each feature point as an individual Gaussian field. Furthermore, we perform experiments by introducing varying levels of noise to the LR image to examine the rationale behind using LR feature values as initial amplitudes. Our findings demonstrate that the final reconstruction results are comparable across different degradation levels; however, the initial convergence rates varied (see Figure \ref{figure 9}). Therefore, by directly utilizing the feature values as initial amplitudes, we can effectively improve the convergence speed. GaussianSR exhibits superior early-stage performance compared to LIIF \cite{LIIF} and ITSRN \cite{ITSRN} approaches, as demonstrated by the results presented in Table 4 of the main article.

\begin{figure}[t]
  \centering
  \includegraphics[width=1\linewidth]{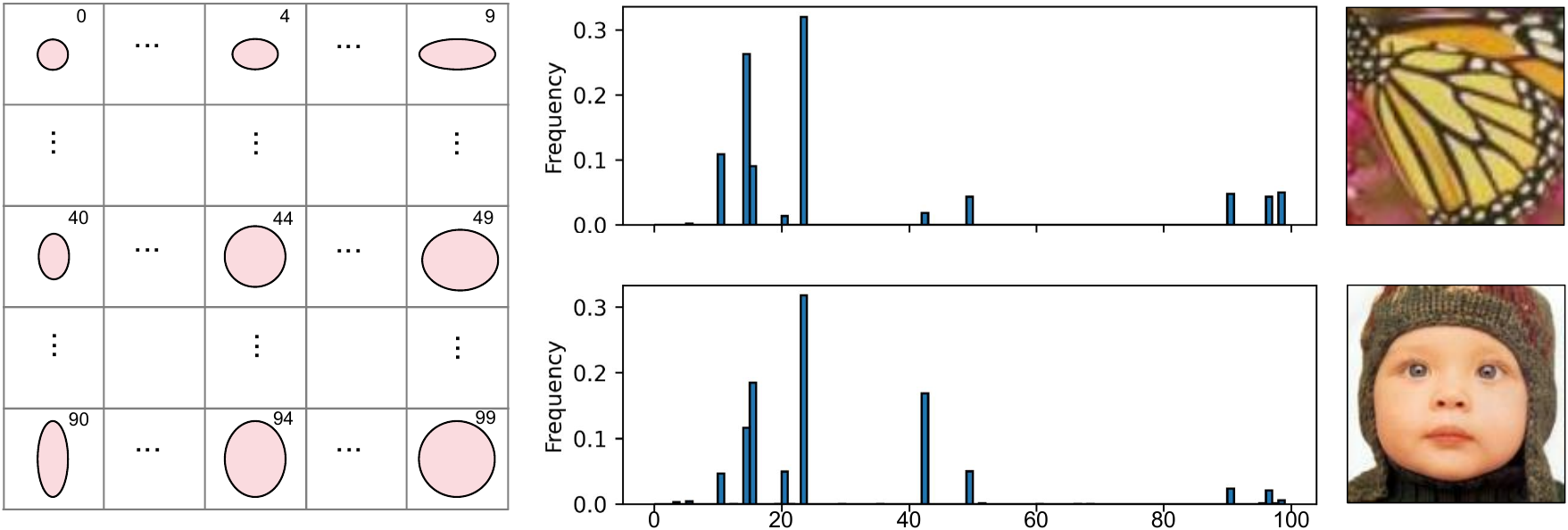}
   \caption{Visualization of Gaussian kernel selection frequencies for 2 example images. The left column displays the Gaussian field index. The middle and right columns are dedicated to the 2 images, each comprising a frequency plot and corresponding image itself.}
   \label{figure 8}
\end{figure}

\textbf{Discussion of LR initialization.} \quad The LR feature initialization process represents the LR image features as initialization points within continuous Gaussian feature fields. The value of the LR feature is used as the initial amplitude of the Gaussian, which ensures that subsequent upsampling and reconstruction processes can effectively leverage the spatial and intensity information from the LR image, ultimately leading to high-quality HR image reconstruction.

\section{2D Gaussian Splatting Pseudocode}

\begin{algorithm}[H]
\caption{A Simple Example of 2D Gaussian Splatting}
\begin{algorithmic}[1]
\Require $\Sigma_x, \Sigma_y, \rho$ (covariance parameters), $\text{coords}$ (point coordinates), $\text{colors}$ (point colors), $\text{image\_size}$
\Ensure $\text{final\_image}$ (rendered image)

\State Compute the covariance matrix $\mathbf{\Sigma}$ using $\Sigma_x, \Sigma_y, \rho$
\State Check if $\mathbf{\Sigma}$ is positive semi-definite
\State Compute the inverse of $\mathbf{\Sigma}$: $\mathbf{\Sigma}^{-1}$
\State Create a 2D grid $\mathbf{x}, \mathbf{y}$ in the range $[-5, 5]$
\State Compute the Gaussian kernel $\mathbf{K}$ using $\mathbf{x}, \mathbf{y}$, and $\Sigma^{-1}$
\State Normalize $\mathbf{K}$ to $[0, 1]$ range
\State Repeat $\mathbf{K}$ along the channel dimension to match colors
\State Pad $\mathbf{K}$ with zeros to match $\text{image\_size}$
\State Create a batch of 2D affine transformation matrices $\Theta$ using $\text{coords}$
\State Apply affine transformations to $\mathbf{K}$ using $\Theta$ to obtain $\mathbf{K}_\text{translated}$
\State Multiply $\mathbf{K}_\text{translated}$ with $\text{colors}$ to get image layers
\State Sum the image layers to obtain $\text{final\_image}$
\State Clamp $\text{final\_image}$ to $[0, 1]$ range
\State Permute $\text{final\_image}$ to match channel order

\Return $\text{final\_image}$
\end{algorithmic}
\end{algorithm}

\begin{figure}[t]
  \centering
  \includegraphics[width=1\linewidth]{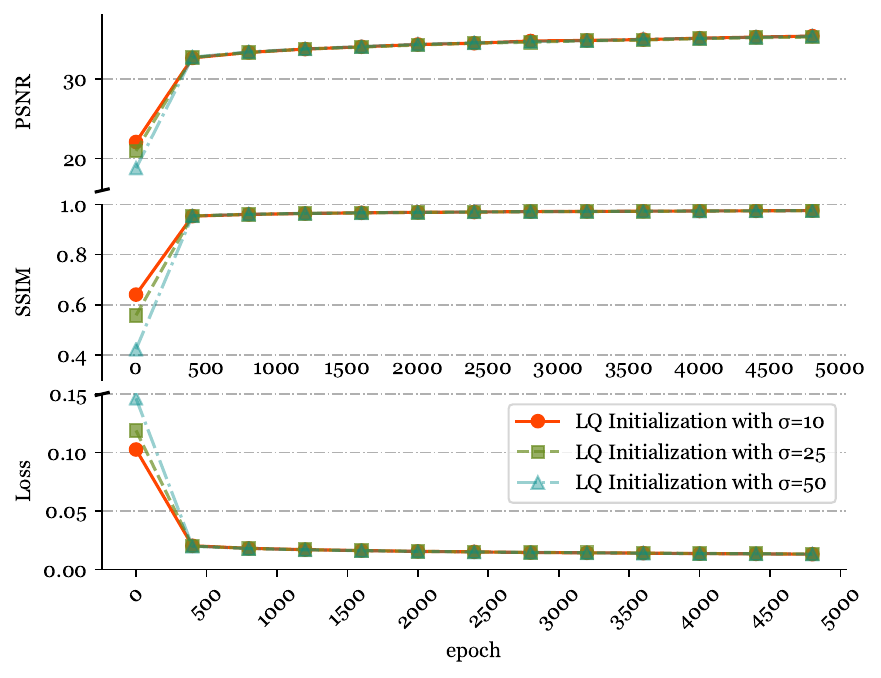}
   \caption{Reconstruction quality with low-resolution feature initialization across different degradation levels.}
   \label{figure 9}
\end{figure}

\section{More Visual Comparison}

\begin{figure}[t]
  \centering
  \includegraphics[width=1\linewidth]{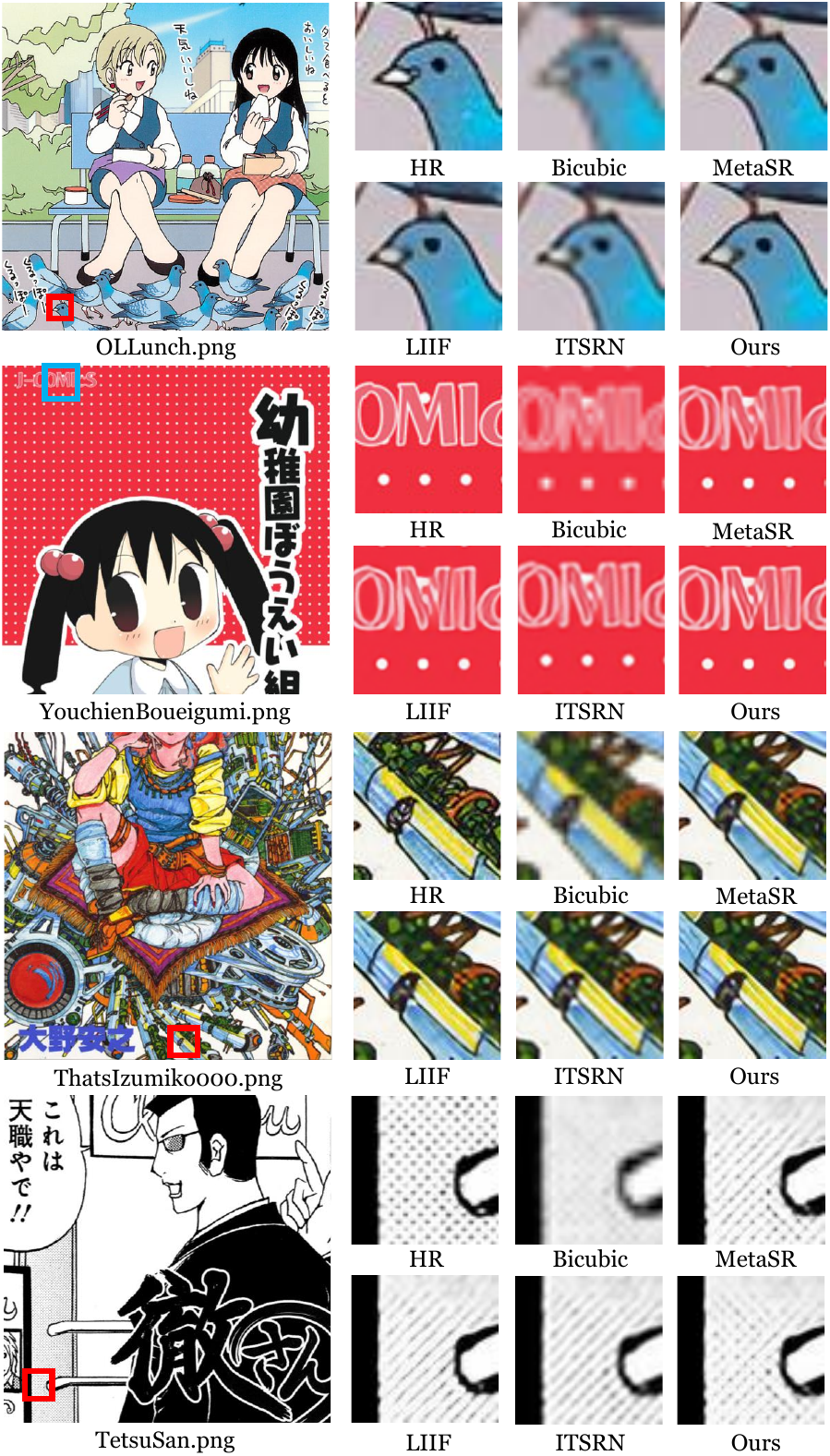}
   \caption{Qualitative comparison for $\times$3 SR to other ASSR methods \cite{MetaSR,LIIF,ITSRN} on Manga109 dataset \cite{Manga109}. EDSR-baseline \cite{EDSR} is used as an encoder for all methods. Zoom in for a better view.}
   \label{figure 10}
\end{figure}

\begin{figure}[t]
  \centering
  \includegraphics[width=1\linewidth]{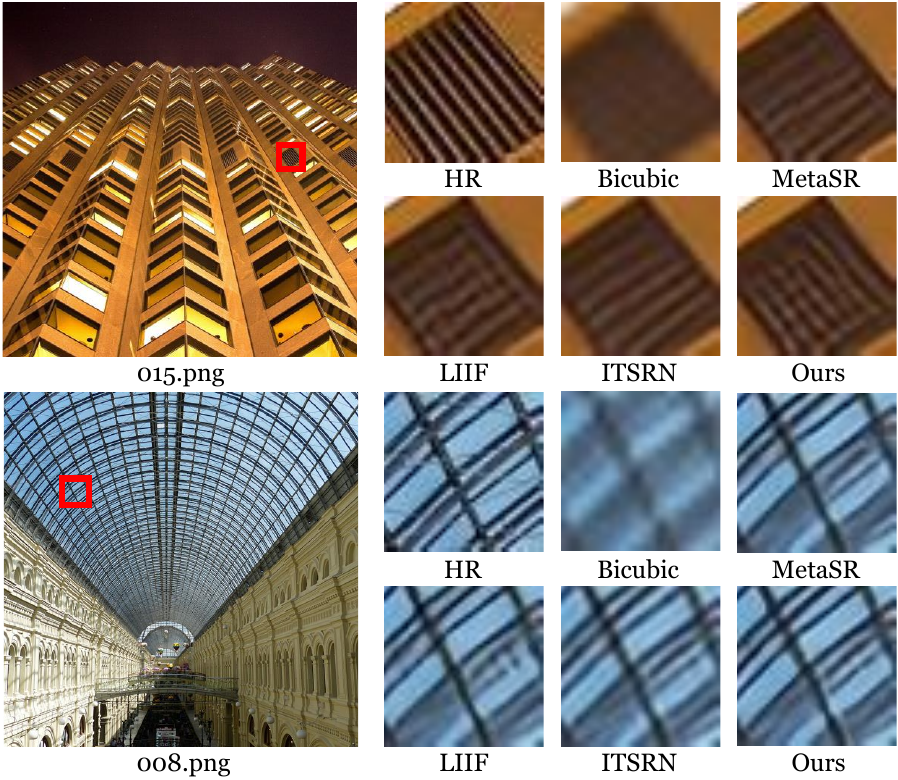}
   \caption{Qualitative comparison with non-integer scale factors $\times$3.3 on Urban100 dataset \cite{urban100}. Zoom in for a better view.}
   \label{figure 11}
\end{figure}

\begin{figure}[t]
  \centering
  \includegraphics[width=1\linewidth]{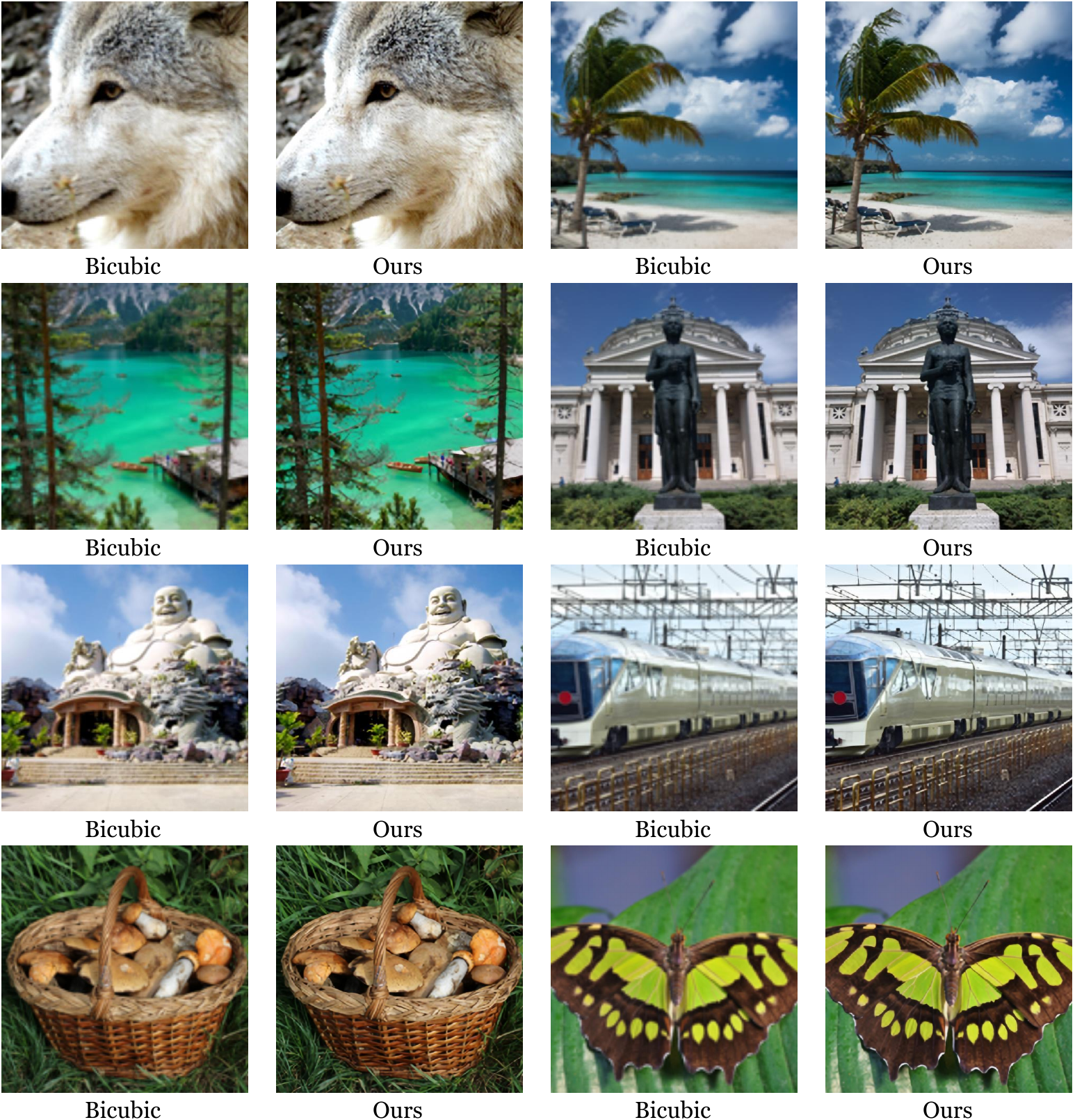}
   \caption{Visual demonstration of GaussianSR for $\times$10 SR. Please zoom in for a better view.}
   \label{figure 12}
\end{figure}

Figure \ref{figure 10} provides a qualitative analysis of the performance of GaussianSR in comparison with other arbitrary-scale super-resolution methods. The evaluation highlights the exceptional capability of our approach in generating high-resolution images with enhanced texture clarity and fidelity. This is particularly evident in the first and last rows of the figure. In the first row, while other methods fail to accurately reconstruct the bird's eyes, our method preserves this critical detail, closely resembling the original image. In the fourth row, which features repetitive and intricate textural structures, our method outperforms others by effectively reconstructing these complex details without introducing artifacts such as texture blending, even when the low-resolution input information is significantly degraded.

Figure \ref{figure 11} further showcases the visual effectiveness of our method in handling non-integer scaling factors. GaussianSR achieves superior restoration results on complex and fine textures, minimizing the presence of artifacts while maintaining a close resemblance to the original image. Furthermore, Figure \ref{figure 12} demonstrates the visual efficacy of our proposed GaussianSR when applied to the challenging task of extreme super-resolution at a scale factor of ×10. Notably, the super-resolved images generated by GaussianSR exhibit significantly sharper and more naturally delineated edges compared to those produced by the conventional bicubic interpolation method. This qualitative assessment highlights the superior performance of our approach in preserving high-frequency details and maintaining the structural integrity of the upscaled images, even under such demanding magnification conditions. The ability of GaussianSR to effectively capture and reconstruct fine textural information while minimizing the introduction of visual artifacts underscores its potential as a state-of-the-art solution for extreme-scale image super-resolution.

\section{Limitation and Future work} 
During the rendering process, for each Gaussian field, we generate a tensor with dimensions [batch size, channel, height, width]. The memory overhead of this five-dimensional vector is substantial. Despite employing techniques such as feature unfolding to reduce memory usage, there remains significant room for improvement. It is noteworthy that this five-dimensional tensor is inherently sparse, as only the neighborhood around the center of each Gaussian field contains values. Therefore, by optimizing matrix multiplication operations on CUDA to better accommodate such sparsity, we can significantly reduce computational speed and memory consumption.

\section{Conclusion}

The proposed GaussianSR pipeline represents a paradigm shift in image super-resolution by introducing a continuous feature field characterized by a Gaussian distribution, departing from the traditional discrete feature storage methodology. This approach enables natural implementation of arbitrary-scale upsampling, as any point within the feature field can be explicitly determined according to the Gaussian distribution. By reconceptualizing pixel representations as continuous Gaussian fields, GaussianSR bridges the gap between discrete feature representations and super-resolution tasks, allowing for a more natural and fluid representation of each pixel with fewer parameters. Furthermore, the classifiers were trained to assign optimal Gaussian kernels to each pixel, effectively tailoring the model to various input characteristics. Experimental results demonstrate that GaussianSR enhances super-resolution performance while reducing computational overhead associated with parameters, highlighting the potential of integrating Gaussian expressions in computer vision tasks traditionally dominated by INR-based methodologies. The proposed framework not only pioneers a novel approach in arbitrary-scale super-resolution but also opens avenues for substantial advancements in how visual information is conceptualized and processed.

\section{Acknowledgements}

This work was partly supported by the National Natural Science Foundation of China (Nos.62171251 $\&$ 62311530100) and the Special Foundations for the Development of Strategic Emerging Industries of Shenzhen (No.KJZD20231023094700001).


{\small
\bibliographystyle{ieee_fullname}
\bibliography{egbib}
}

\end{document}